\documentclass{article}

\usepackage{microtype}
\usepackage{graphicx}
\usepackage{booktabs} 

\usepackage{hyperref}

\usepackage[accepted]{icml2025}

\usepackage{amsmath}
\usepackage{amssymb}
\usepackage{mathtools}
\usepackage{amsthm}

\usepackage{subcaption}
\usepackage{newfloat}
\usepackage{listings}
\usepackage{multirow}
\usepackage{xcolor}

\usepackage[capitalize,noabbrev]{cleveref}

\theoremstyle{plain}

\theoremstyle{definition}

\theoremstyle{remark}

\usepackage[textsize=tiny]{todonotes}

\icmltitlerunning{Panoptic Diffusion Models: co-generation of images and segmentation maps}

\begin{document}

\twocolumn[
\icmltitle{Panoptic Diffusion Models: co-generation of images and segmentation maps}




\begin{icmlauthorlist}
\icmlauthor{Yinghan Long}{purdue}
\icmlauthor{Kaushik Roy}{purdue}
\end{icmlauthorlist}

\icmlaffiliation{purdue}{Purdue University}

\icmlcorrespondingauthor{Yinghan Long}{    long273@purdue.edu}
\icmlcorrespondingauthor{Kaushik Roy}{kaushik@purdue.edu}

\icmlkeywords{Machine Learning, ICML}

\vskip 0.3in
]



\printAffiliationsAndNotice{} 


\begin{abstract}
 Recently, diffusion models have demonstrated impressive capabilities in text-guided and image-conditioned image generation. However, existing diffusion models cannot simultaneously generate an image and a panoptic segmentation of objects and stuff from the prompt.  Incorporating an inherent understanding of shapes and scene layouts can improve the creativity and realism of diffusion models. To address this limitation, we present Panoptic Diffusion Model (PDM), the first model designed to generate both images and panoptic segmentation maps concurrently. PDM bridges the gap between image and text by constructing segmentation layouts that provide detailed, built-in guidance throughout the generation process. This ensures the inclusion of categories mentioned in text prompts and enriches the diversity of segments within the background. We demonstrate the effectiveness of PDM across two architectures: a unified diffusion transformer and a two-stream transformer with a pretrained backbone. We propose a Multi-Scale Patching
mechanism to generate high-resolution segmentation maps. Additionally, when ground-truth maps are available, PDM can function as a text-guided image-to-image generation model. Finally, we propose a novel metric for evaluating the quality of generated maps and show that PDM achieves state-of-the-art results in image generation with implicit scene control.
\end{abstract}

%

\section{Introduction}
\label{sec:intro}

\begin{figure}[tbp]
\begin{subfigure}[h]{0.3\linewidth}
\includegraphics[width=\linewidth]{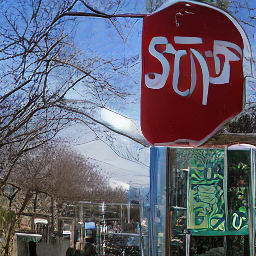}
\caption{``An upside down stop sign by the road."}
\end{subfigure}
\hfill
\begin{subfigure}[h]{0.3\linewidth}
\includegraphics[width=\linewidth]{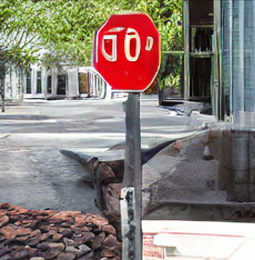}
\caption{PDM generated image of a stop sign.
}
\end{subfigure}%
\hfill
\begin{subfigure}[h]{0.3\linewidth}
\includegraphics[width=\linewidth]{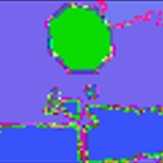}
\caption{PDM generated octagon mask for the stop sign.}
\end{subfigure}%
\\\begin{subfigure}[h]{0.3\linewidth}
\includegraphics[width=\linewidth]{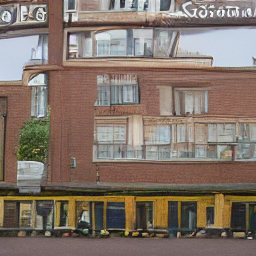}
\caption{``A fire hydrant on the side of the street.
"}
\end{subfigure}
\hfill
\begin{subfigure}[h]{0.3\linewidth}
\includegraphics[width=\linewidth]{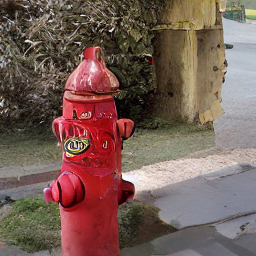}
\caption{PDM generated image of a fire hydrant.\\
}
\end{subfigure}%
\hfill
\begin{subfigure}[h]{0.3\linewidth}
\includegraphics[width=\linewidth]{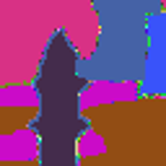}
\caption{PDM generated mask for a fire hydrant.}
\end{subfigure}%
\\
\begin{subfigure}[h]{0.3\linewidth}
\includegraphics[width=\linewidth]{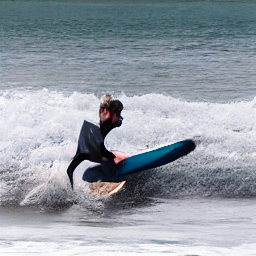}
\caption{``A man with a wet suit on standing on a surfboard in the water."}
\end{subfigure}
\hfill
\begin{subfigure}[h]{0.3\linewidth}
\includegraphics[width=\linewidth]{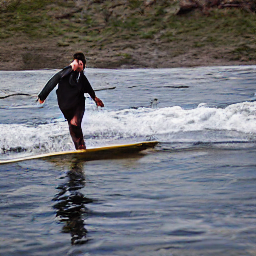}
\caption{PDM generated image of a man surfing in the water.\\
}
\end{subfigure}%
\hfill
\begin{subfigure}[h]{0.3\linewidth}
\includegraphics[width=\linewidth]{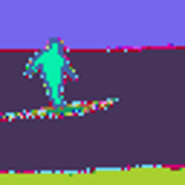}
\caption{PDM generated masks for person, sky, and sea\\}
\end{subfigure}%
\\
\begin{subfigure}[h]{0.3\linewidth}
\includegraphics[width=\linewidth]{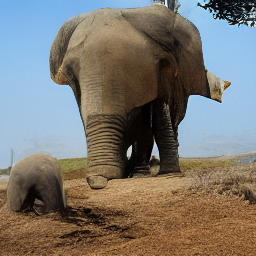}
\caption{``Several elephants walking together in a line near water."}
\end{subfigure}
\hfill
\begin{subfigure}[h]{0.3\linewidth}
\includegraphics[width=\linewidth]{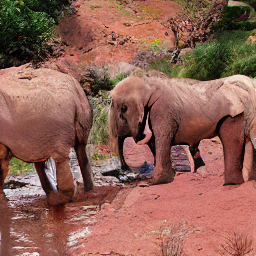}
\caption{PDM generated image of several elephants near a river.
}
\end{subfigure}%
\hfill
\begin{subfigure}[h]{0.3\linewidth}
\includegraphics[width=\linewidth]{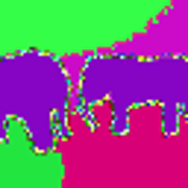}
\caption{PDM generated masks for elephants, river, grass, and sand.}
\end{subfigure}%
\caption{Left: images generated by a regular diffusion model (U-ViT) based on the text prompt. Right: images and masks generated by a Panoptic Diffusion Model based on the same text.}
\label{fig:compare}
\end{figure}

Diffusion models have recently outperformed other generative models, demonstrating a strong ability to generate high-quality, photorealistic images and creative videos with high fidelity \cite{diffusion-beat-gan, imagen, dalle, stablediffusion, glide, sora2024, ho2022imagenvideo, ho2022video, bartal2024lumiere, singer2022makeavideotexttovideogenerationtextvideo}. Their success has drawn significant attention to generative AI, marking it as the next frontier following the achievements of AI in classification tasks.
However, text-guided image generation often lacks control over the spatial structure of the image \cite{zhang2023controlnet}. Current diffusion models have difficulty understanding shapes of objects because the diffusion process is uniformly applied to every pixel, without regard to the segment it belongs to. As a result, they may generate objects with unrealistic shapes and miss components mentioned in the text, leading to images that are perceived as artificial, as shown in the left column of Fig.\ref{fig:compare}.


To address this issue, we propose teaching diffusion models to understand object shapes and scene structures through panoptic segmentation, which provides information about both countable objects in the foreground and background elements that complements text prompts \cite{panoptic}. Recent works, such as ControlNet, have demonstrated that using images with complex layouts as conditions, in addition to text prompts, can precisely control the generation process \cite{zhang2023controlnet}. These studies show that image-guided generation can better align with users' specific imaginings expressed through both text and image prompts. Inspired by this, we anticipate that if diffusion models generate segmentation maps alongside images to provide inherent guidance, they can utilize spatial composition information to create more realistic images. 


The co-generation of images and masks is nontrivial and challenging because it represents a \textbf{dual} problem. Unlike previous approaches that rely on either a clean image or a segmentation map as a stable condition to generate the other, our model tackles the complex task of simultaneously denoising both an image and its corresponding map \cite{zhang2023controlnet, Pix2seq-diffusion}. To address this, we designed a new paradigm to solve the dual diffusion problem.
Compared to using predefined segmentation maps, co-generation preserves the diversity and flexibility of the images. By generating panoptic segmentation maps, Panoptic Diffusion Models (PDMs) provide intrinsic control over image generation, while the images in turn ensure that the map generation remains coherent. Since the generation of both segmentation maps and images is guided by text, the model learns the correlation between text, images, and maps. With its enhanced scene understanding capabilities, PDMs represent a significant step towards photorealistic image generation.  

We design both a one-stream PDM and a two-stream model that incorporates a pretrained image generation stream. For training the two-stream model, we fix the image stream and efficiently fine-tune the segmentation stream.
Compared to using two separate models in a sequence for generating segmentation maps and images, a unified model is more efficient and advantageous due to its ability of supervised learning between segmentation and images. To reduce the computation overhead, we propose a Multi-Scale Patching mechanism to directly generate high-resolution segmentation maps, instead of processing the latent by a VAE decoder.
The pixel-level segmentation maps generated by PDM can benefit downstream computer vision tasks, such as autonomous driving.

The major contributions are listed below:

1. We propose a unified diffusion model that generates both images and panoptic segmentation maps. This model inherently understands scene structures through collaborative training with multimodal data, requiring no priors and providing self-control.

2. We adapt the fast ODE solver for image denoising to facilitate simultaneous image and map generation. The iterative denoising of images and maps is interlinked, ensuring consistency between them.

3. We develop a two-stream diffusion model and apply efficient fine-tuning techniques. This approach leverages pretrained diffusion models and extends their capabilities by incorporating segmentation maps.

4. By multi-scale patching, PDM generates segmentation maps that scale up to four times the latent size without requiring a super-resolution model. We also introduce a new metric for evaluating the quality of the generated maps.

\begin{figure*}[htbp]
    \centering
    \includegraphics[width=0.9\linewidth]{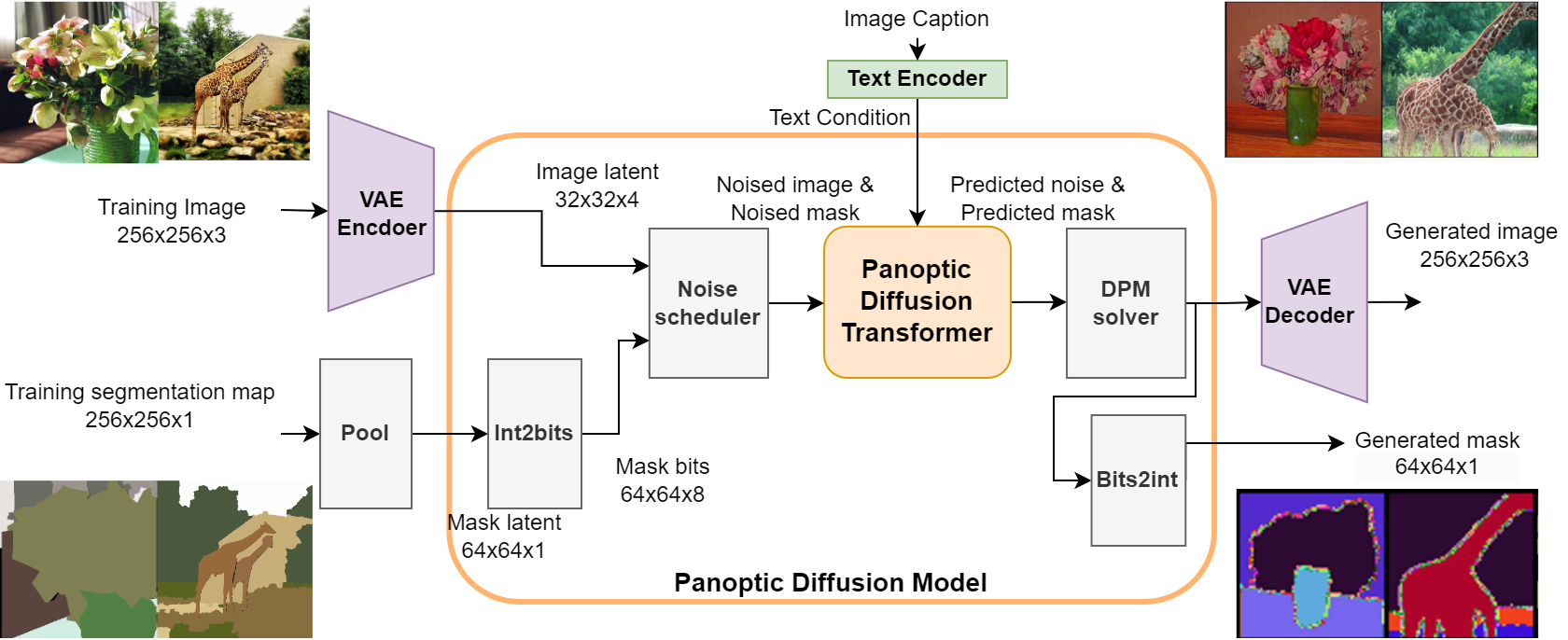}
    \caption{Pipeline of Panoptic Diffusion Models}
    \label{fig:pipeline}
\end{figure*}

\section{Related works}

\subsection{Diffusion Models for Image Generation}
Denoising Diffusion Probabilistic Models (DDPM) use a Markov chain to gradually add scheduled noises to images in the forward process and then parameterize the transition by a neural network trained to predict the noise \cite{ho2020denoising}. During inference, a diffusion model starts from random noise and gradually reverses it to reconstruct the image.
A well-known drawback of diffusion models is that they require a large number of steps to generate samples iteratively. To improve efficiency, researchers have proposed various modifications to diffusion models \cite{improvedDDPM}. DDIM demonstrates that diffusion models can operate in a non-Markovian manner, resulting in shorter generative chains \cite{ddim}. Additionally, distillation algorithms have been introduced to further accelerate the multi-step inference process\cite{progressive-diffusion, berthelot2023tract, ren2024hypersd}. We use a fast solver for our panoptic diffusion model, which is a modified version of DPM Solver++ that can solve the reverse of the diffusion process in 10-50 steps \cite{lu2023dpmsolverfastsolverguided, lu2022dpm}.

The backbone neural network for a diffusion model is typically a UNet, which is composed of convolutional layers and attention blocks, or a diffusion transformer that relies solely on attention mechanisms \cite{stablediffusion, Peebles2022DiT}. Another variant, UViT, is a type of diffusion transformer that retains skip connections, allowing later layers to access information from earlier layers, thereby enhancing alignment \cite{uvit}.

There are three main methods for applying conditions to a diffusion model. The first approach, used in stable diffusion, involves cross-attention between the image and the conditions \cite{stablediffusion}. The second method appends condition embeddings as tokens to the image patches \cite{uvit}. The third approach uses an adaptive norm layer to integrate conditions with the hidden states \cite{Peebles2022DiT}. In our panoptic diffusion models, we opt for the second method because the transformer can leverage self-attention to learn the relationships between images and maps, treating them as conditions for each other.
 During inference, we apply classifier-free guidance similar to \citet{glide} and \citet{ho2022classifierfree}.

\subsection{Image Segmentation}
Object detection requires generating bounding boxes and fine-grained masks, tasks traditionally accomplished by convolutional neural networks such as Fast R-CNN \cite{fastRCNN} and Mask R-CNN \cite{maskRCNN}. In \citet{DETR}, researchers introduced the use of transformers to generate binary masks by inputting object queries. Building on this, \citet{cheng2021mask2former} proposed a collaboration between an image encoder backbone and a masked transformer to generate masks, where masked attention replaces cross attention. With advanced segmentation models like Segment Anything \cite{kirillov2023segany} easily segmenting images, segmentation maps hold potential as alternative or complementary training data for image generation tasks.  

Recently, there has been growing interest in applying diffusion models to segmentation masks. For example, \citet{baranchuk2021labelefficient} suggest that the intermediate features of diffusion models can capture semantic information useful for label-efficient segmentation. Similarly, DiffuMask \cite{wu2024diffumasksynthesizingimagespixellevel} and Dataset Diffusion \cite{nguyen2023datasetdiffusiondiffusionbasedsynthetic} generate a synthetic pair of an image and a corresponding segmentation annotation of objects using attention maps. However, directly extracting masks from attention maps lacks the ability to control the generated image in return. Unified diffusion models for image generation and segmentation has shown a potential to refine image generation, such as UniGS \cite{qi2023unigsunifiedrepresentationimage}. While the existing works focuses on semantic segmentation, our method extends to panoptic segmentation, providing both instance and semantic information. This is a crucial distinction and expands the potential applications of our model. 

On the other hand, some previous studies use diffusion models for panoptic segmentation based on given images. In \citet{Pix2seq-diffusion}, a diffusion model comprising an image encoder and a mask decoder is used to extract image features and apply cross attention between these features and the masks. To address the challenge of handling discrete data with diffusion models, \citet{chen2022analog} proposed converting panoptic masks into analog bits during preprocessing. Our approach extends the ability of the diffusion model to co-generate pixel-level panoptic segmentation maps and images, allowing them to influence and control each other.


\subsection{Image Guided Image Generation}
Image guided image generation enables more precise control over the structure of the image and ensures faithfulness to users' illustrative inputs. The input for guidance can have various forms, such as segmentation maps and layouts \cite{stablediffusion,zhang2023controlnet}. Stochastic Differential Editing (SDEdit) perturbs user inputs with Gaussian noises and then synthesizes images by reversing SDE \cite{meng2022sdedit}. They show that when the reverse SDE is not solved from the ending point but a particular timestep, the generated images can achieve a good balance between faithfulness and realism.
Make-a-scene introduces scene-based conditioning for image generation by optionally providing tokens from segmentation maps \cite{gafni2022makeascene}, but this method heavily relies on explicit strategies for tackling panoptic, human, and face semantics. SpaText \cite{spatext2023} employs CLIP \cite{radford2021clip} to convert local text prompts that describe segments into image space and concatenate to the channel dimension of noises. ControlNet can accept user inputs such as canny edges and segmentation masks for conditional control of image generation \cite{zhang2023controlnet}. Prompt-to-prompt image editing controls the generation by cross-attention to ensure similarity between images generated from similar prompts \cite{prompt2prompt}. InstructPix2Pix combines Prompt-to-prompt method with stable diffusion to generate pairs of images from pairs of captions for training, then train the model to modify image pixels following the instructions \cite{brooks2023instructpix2pix}.

These approaches demonstrate that providing various forms of guidance can more accurately control the structure of generated images. Building on this insight, our method assumes that such guidance is crucial for enhancing image quality. Additionally, panoptic diffusion models inherently generate segmentation maps alongside images, offering built-in guidance without the need for additional user input beyond the text prompt.

\subsection{Efficient Deep Learning}
To reduce the number of trained parameters or adapt the model to a new domain, previous works have designed adaptive blocks to fine-tune convolutional neural networks or transformers \cite{adapter, Complexity-adaptive, mou2023t2iadapterlearningadaptersdig}. In our two-stream panoptic diffusion model, the map stream functions similarly to an adapter. To prevent any negative impact on the pretrained weights, we employ zero-initialized convolutional blocks as proposed in \citet{zhang2023controlnet}.

Unlike other works that introduce significant computational overhead to generate segmentation with images, our method maintains the efficiency by leveraging a bit encoding scheme and multi-scale patching. This allows for parallel generation of images and masks without substantial additional computational cost. We will include a comparison of the number of parameters to highlight this advantage.


\section{Panoptic Diffusion}

\subsection{Preprocessing and Postprocessing of Segmentation Maps}
As shown in Fig. \ref{fig:pipeline},we process the panoptic segmentation maps through several steps before feeding them into the diffusion model. Instead of using a binary mask for each object, we load pixel-level panoptic annotations. In a segmentation map $M_0$, each pixel's value is set to the corresponding category ID if it belongs to a segment; otherwise, its value is zero. We then convert these pixel values into analog bits \cite{chen2022analog}. Analog bits are necessary because a standard diffusion model can only generate continuous data, while segmentation classes are discrete and categorical. Since the range of category ID is from 1 to 200, each pixel is represented by 8 binary bits. Prior to noise scheduling, these bits are scaled to the range $[-1,1]$, matching the range of the latent input to the diffusion model. To ensure that the noise can effectively flip the bits, its absolute value must exceed one. Therefore, we set the noise added to the maps as $\epsilon_M\sim\mathcal{N}(0,2*\mathbf{I})$. 

Latent diffusion models use latent representations of images encoded by a variational autoencoder(VAE) as inputs. However, using a separate VAE for encoding and decoding high-resolution segmentation maps is inefficient. We address this issue by pooling and multi-scale patching. To achieve high-resolution maps and enable more precise control, we first pool the maps to match one, two, or four times the height and width of the image latents. We use min pooling to prioritize smaller category numbers, as the COCO dataset annotations categorize 1-91 as thing categories and 92-200 as stuff categories. Next, we set the patch size of the maps to be one, two, or four times that of the images. This approach ensures that, after patchifying, the sizes of the image and map features align. Given that images have three RGB channels while maps have only one channel for the category ID before preprocessing, using a larger patch size is effective for extracting hidden features from segmentation maps. Consequently, this method allows us to generate higher-resolution maps without the need for an additional VAE or a larger latent size.

For postprocessing, the output values predicted by the diffusion model are thresholded at zero. Negative values are treated as zero bits, while positive values are considered one bits. Subsequently, these output bits are converted back into category numbers.

\subsection{Forward Diffusion Process}
In the forward pass of the diffusion process \cite{ho2020denoising}, random noise $\epsilon\sim\mathcal{N}(0,\mathbf{I})$ is added to the image latent $x_0$ according to the noise scheduler. With a total of $n$ steps, each step updates the noisy image $x_t$ from the previous step $x_{t-1}$, using scaling factors $\alpha$ and $\beta$ provided by the noise scheduler. This process forms a Markov chain. Consequently, the noisy image $x_t$ can be simplified and calculated directly from $x_0$.
\begin{align}
    x_{t} = \sqrt{\alpha_{t}} \cdot x_{t-1} + \beta_{t}\epsilon  \label{addnoise}\\
    x_{t} = \sqrt{\Bar{\alpha}} \cdot x_0 + \sigma_t \epsilon 
\end{align}
where $\alpha_t$ are close to 1 and $\beta_t=1-\alpha_t$. The cumulative factor $\Bar{\alpha}=\prod_{i=1}^t\alpha_{i}$, and the noise is scaled by $\sigma_t=\sqrt{1-\Bar{\alpha}}$.  

To learn to denoise panoptic segmentation maps, we create another random Gaussian noise $\epsilon_M\sim\mathcal{N}(0,2*\mathbf{I})$ and add it to the ground-truth maps $M_0$. The same noise scheduler is used to add noises to maps.
\begin{align}
    M_{t} = \sqrt{\Bar{\alpha}} \cdot M_0 + \sigma_t \epsilon_M  \label{eq.mt}
\end{align}
where $M_t$ is the noised map at timestep $t$. \\

\subsection{Reverse Diffusion Process}
 The panoptic diffusion model outputs $\epsilon_\theta$, which estimates the noise $\epsilon$. Using this estimated noise, we compute the predicted image $\Tilde{x_0}$. When incorporating the map as an additional input to the diffusion model, the equation for predicting the image is given by Eq. \ref{predictx0}. To accelerate inference, we utilize a fast DPM solver to compute $x_{t_{i-1}}$ from $x_{t_i}$ \cite{lu2022dpm, lu2023dpmsolverfastsolverguided}. By using discontinuous time steps $t_{i}$ and $t_{i-1}$, this method can skip intermediate steps, reducing the total number of sampling steps required. The first-order solver is described in Equation \ref{dpm}, where $h_i$ represents the difference in the log signal-to-noise ratio between different steps ($h_i= \log(\alpha_{t_i}/\sigma_{t_i})-\log(\alpha_{t_{i-1}}/\sigma_{t_{i-1}})$). Details on a third-order solver can be found in Appendix \ref{appen.dpm}. 

\begin{align}
    \Tilde{x_0} (x_{t_i}, M_{t_i}, C, t_i)= \dfrac{x_{t_i}- \sigma_t \epsilon_\theta(x_{t_i},M_{t_i},C, t_i)}{\sqrt{\Bar{\alpha}}}  \label{predictx0}\\
    x_{t_{i-1}}= \dfrac{\sigma_{t_{i-1}}}{\sigma_{t_i}} x_{t_i} - \alpha_{t_i}(e^{-h_i}-1) \Tilde{x_0} (x_{t_i}, M_{t_i}, C, t_i)
    \label{dpm}
\end{align}

The other output of a panoptic diffusion model is $M_\theta$, which is a prediction of $M_0$.  Drawing inspiration from DPM-solver++, we use the following equation to estimate $M_{t_{i-1}}$ from the previous step. It is important to note that the model directly estimates $M_0$ rather than the noise added to the segmentation map, as predicting $\epsilon_M$ does not provide effective guidance for the images. By training the diffusion model with panoptic segmentation maps, it incorporates intrinsic self-control into the image generation process.
\begin{align*}
    M_{t_{i-1}}= \dfrac{\sigma_{t_{i-1}}}{\sigma_{t_i}} M_{t_i} - \alpha_{t_i}(e^{-h_i}-1) M_{\theta} (x_{t_i}, M_{t_i}, C, t_i)
\end{align*}

In a special case where ground truth maps are provided as conditions, the diffusion model will focus solely on predicting the images. This allows users to have customized control for generating desired images, similar to existing methods \cite{zhang2023controlnet}. However, this approach limits the diversity of the generated images.

Since the generation of $x_{t-1}$ and $M_{t-1}$ relies on $x_{t}$ and $M_{t}$, they form a dual problem. Improvements in the quality of the generated masks and images influence each other. Consequently, according to the scaling law, a larger diffusion model can produce more accurate masks, which in turn provides better control and further enhances image quality.



\subsection{Dual training and generation}
Let the inputs to a panoptic diffusion model at each timestep be image latent $x_t$, mask $M_t$, text condition encoded by a text encoder $C$, and timestep $t$. The conditional probability of $x_{t-1}$ and $M_{0}$ is given by
\begin{align}
 &    P(x_{t-1},  M_{0}| x_t, M_t, c)\nonumber\\& =  P(x_{t-1} | x_t, M_t, M_{0}, c) \cdot  P( M_{0}| x_t, M_t, c) \label{eq.cond} 
\end{align}
Equation \ref{eq.cond} show that it is feasible to predict the segmantation map $M_0$ first, then use it as a condition to predict $x_{t-1}$. However, when using a unified model to predict both $x_{t-1}$ and $M_{0}$, the intermediate features already contain the segmentation information used to predict $M_{0}$. Through self-attention, the map features can inherently condition $x_{t-1}$. Therefore, it is reasonable to predict $x_{t-1}$ and $M_{0}$ simultaneously. By taking the logarithm of the probability, we can optimize the model by combining the losses associated with image denoising and segmentation map generation.
\begin{align}
&  \log P(x_{t-1}, M_{0}| x_t, M_t, c) \nonumber \\ & =  \log P(x_{t-1} | x_t, M_t, M_{0}, c) +  \log P(M_{0}| x_t, M_t, c)
\end{align}


The training algorithm is outlined in Algorithm~\ref{algo.1}. We use Mean Squared Error (MSE) loss to optimize the predicted noises for both image and segmentation map denoising. Specifically, the target for image denoising is the noise $\epsilon$, while the target for mask generation is the ground-truth $M_0$. The losses for images and maps are summed to perform gradient backpropagation. During inference, the diffusion model iteratively denoises both images and maps, as detailed in Algorithm \ref{algo.2}.

\subsubsection{Classifier-free Map Guidance}
Classifier-free diffusion guidance was introduced to balance sample quality and diversity without relying on a classifier \cite{ho2022classifierfree}. This approach involves alternating between an unconditional and a conditional diffusion model during training, and using a weighted sum of the results from both models during inference. For panoptic diffusion models, we only remove the text conditions while keeping the map conditions active. Specifically, we set the context condition to empty text with a probability of 0.1 during training ($C=\varnothing$). When the context is empty, the diffusion model is guided solely by the bidirectional control between images and segmentation maps. Let $\theta_1$ represent the output with regular conditioning and $\theta_2$ represent the output with empty text. During inference, these outputs are weighted by $\gamma$, which is set to 1.0 by default.
\begin{align*}
    \epsilon_\theta = \epsilon_{\theta1} + \gamma (\epsilon_{\theta1}-\epsilon_{\theta2}); \quad
    M_\theta = M_{\theta1} + \gamma (M_{\theta1}-M_{\theta 2})
\end{align*}

\begin{algorithm}[h]
 \caption{Training of Panoptic Diffusion model \label{algo.1}}
\begin{algorithmic}
   \STATE {\bfseries Input:} Ground truth Masks $M_0$; Images $x_0$; Text condition $C$; Total number of steps $T$
\STATE {\bfseries Output:} Predicted noise $\epsilon_{\theta}$, Predicted mask $M_{\theta}$ 
\STATE $\epsilon$ = normal(mean=0, std=1) \\
\STATE $\epsilon_m$ = normal(mean=0, std=2) \\
\STATE $M_0$ = int2bits($M_0$)\\
\STATE t = randn(1,T)\\
\STATE $x_t$ = scheduler($x_0$, $\epsilon$, t) \\
\STATE $M_t$ = scheduler($M_0$, $\epsilon_m$ , t) \\
\STATE $\epsilon_{\theta}$,  $M_{\theta}$ = DiffusionModel($x_t$, $M_t$, $C$, $t$) \\
\STATE $loss_{x}$ = MSE($\epsilon$, $\epsilon_{\theta}$) \\
\STATE $loss_{m}$ = MSE($M_0$, $M_{\theta}$)\\
\STATE $loss= loss_{x} + loss_{m}$
 \end{algorithmic}
\end{algorithm}

\begin{algorithm}[h]
 \caption{Inference of Panoptic Diffusion model using DPM solver \label{algo.2}}
\begin{algorithmic}
   \STATE {\bfseries Input:} Text $C$; Total number of steps $T$
   \STATE {\bfseries Output:}Generated image $x_0$, Generated mask $M_0$
   \STATE $x_t$= normal(mean=0, std=1) \\
   \STATE $M_t$ = normal(mean=0, std=1) \\
   \STATE Sample a set of steps $T$ from n to 0 \\
    \FOR{$t$ in $T$} 
    \STATE \# Run the diffusion model\\
    \STATE $\epsilon_{\theta}$, $M_{\theta}$ = DiffusionModel($x_t$, $M_t$, $C$,$t$)\\
    \STATE \# Update predicted images and masks\\
    \STATE $x_0=\dfrac{x_t- \sigma_t \epsilon_{\theta}}{\sqrt{\Bar{\alpha}}}$\\
    \STATE $x_{t}$, $M_{t}$ = dpmSolver($x_0$, $M_{\theta}$,$X_t$, $M_t$,$t$)\\
    \ENDFOR
\end{algorithmic}
\end{algorithm}

\subsection{Architecture of Panoptic Diffusion Models}
\subsubsection{One-stream Panoptic Diffusion Models}
We first modify a U-ViT to a panoptic diffusion model \cite{uvit}. We start by patchifying the map input $M_t$ using a convolutional layer and adding positional embeddings. These map embeddings are then concatenated with the image, text, and time embeddings and processed through attention blocks. Since U-ViT treats all inputs as tokens and applies self-attention among them, the segmentation maps can be treated as tokens in the same manner. At the end of the transformer, we separate the features related to images and segmentation maps, using distinct convolutional layers to unpatchify and predict the outputs.

In the special case that the ground truth maps are provided, only the loss of images will be used for optimization. To ensure that map features are included in the gradient backpropagation, they are added to the image features before the final output convolutional layer.

\subsubsection{Two-stream Panoptic Diffusion Models}
\begin{figure}[hbtp]
    \centering
    \includegraphics[width=0.7\linewidth]{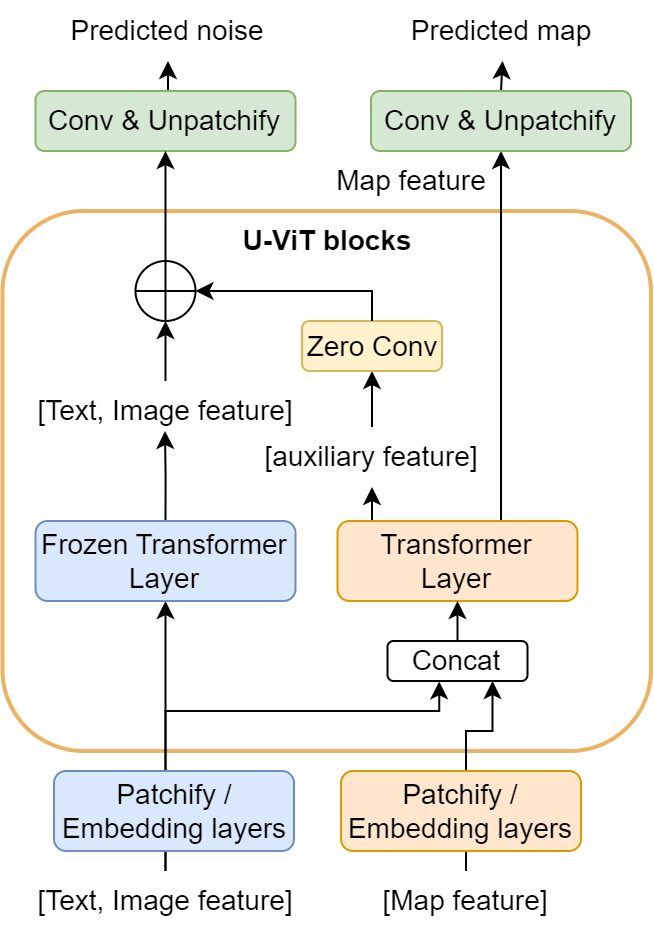}
    \caption{Two-stream panoptic diffusion model. There are a pretrained image stream on the left and a fine-tuned segmentation map stream on the right.}
    \label{fig:two-stream}
\end{figure}

To leverage a pretrained model as the backbone, we design a two-stream diffusion model consisting of a pretrained image stream and a segmentation map stream, as illustrated in Fig. \ref{fig:two-stream}. During fine-tuning, the transformer layers of the image stream are kept frozen while the map stream is adjusted. The map stream processes image features and conditions from the previous block, then concatenates them with map features. Through self-attention, the map features and image features become interrelated within the map stream. The auxiliary image feature output from the map stream is added back to the image stream via a zero-convolution layer. This setup ensures specific control over the image stream and allows gradients to be backpropagated from the loss of image generation. The zero-convolution layer has zero initial weights and no bias \cite{zhang2023controlnet}. Unlike ControlNet, which uses only the encoder part of the map stream to generate control signals, our model employs encoder-decoder U-shaped transformers in both streams to co-generate images and segmentation maps.

\subsection{Evaluation metric for generated maps}
We propose a new metric to evaluate the quality of generated segmentation maps by measuring the difference in the number of pixels labeled as each category. While Panoptic Quality \cite{panoptic} uses Intersection over Union (IoU) to assess segmentation maps, this metric is not suitable to evaluate maps co-generated with images.
We introduce the Mean Count Difference (MCD) metric. MCD evaluates the quality of generated maps by counting the frequency $f$ of each category in both the ground-truth and generated maps, then summing their absolute differences. This sum is divided by the total number of pixels, calculated as the product of the height and width. Given that object locations on the generated map are not fixed, comparing category frequencies rather than direct pixel values provides a more meaningful assessment. The metric ranges from $[0, 2]$, where zero indicates identical segmentation maps and larger values indicate greater differences.
\begin{align*}
    f= bincount(M_0); \quad f'=bincount(M_{\theta})\\
    MCD= \dfrac{\sum(|f-f'|)}{H*W}
\end{align*}

\section{Experiments}
We train our model using the COCO2017 dataset \cite{lin2015microsoftcococommonobjects}, which includes both panoptic segmentation maps and image captions. The COCO2017 dataset comprises 118k training samples and 5k validation samples. Images are projected into latent space using a VAE model provided by Stable Diffusion \cite{stablediffusion, VQVAE}, while text conditions are encoded using the CLIP encoder from OpenAI (clip-vit-large-patch14) \cite{radford2021clip}. We implement both one-stream and two-stream panoptic diffusion models (PDM) based on U-ViT \cite{uvit}. In contrast to commercial models with billions of parameters, our models are significantly smaller. The one-stream PDM has 45 million parameters, while the two-stream PDM has 95 million parameters. The image latent size is $32 \times 32 \times 4$, with a height and width of 32 and a latent channel count of 4. The segmentation map’s height and width can be 32, 64, or 128, depending on the patch factor, and it has 8 channels, representing 8 analog bits after conversion. The diffusion model’s output image latents are decoded by a VAE decoder to produce $256 \times 256$ images.

\begin{table}[htbp]
    \centering
    \begin{tabular}{|c |c c | }
    \hline
    Model & FID($\downarrow$) & CLIP($\uparrow$) \\
    \hline
   GLIDE \cite{glide} & 12.24 & $\sim$28 \\
    \hline
   Imagen  \cite{imagen} & 7.27 & $\sim$27 \\
    \hline
   VQ-Diffusion \cite{VQVAE}   & 13.86 &   -\\
    \hline
    UViT \cite{uvit} & 8.29 & 27.37 \\
    \hline
      One-stream PDM & 18.52 & 26.32 \\
     \hline
       Two-stream PDM   &  10.99& 27.53 \\
    \hline
        One-stream PDM given maps  & 8.21 & 28.40 \\
        \hline
        Two-stream PDM given maps  & 11.61 & 28.19 \\
    \hline
    \end{tabular}
    \caption{Quantitative Evaluation Results of COCO dataset. }
    \label{tab:results}
\end{table}

\begin{table}[hbt]
    \centering
    \begin{tabular}{c|c|c|c|c}
       \hline
     Model & FID($\downarrow$) & CLIP($\uparrow$) & Patch & MCD\\
    \hline
      One-stream PDM & 18.52 & 26.32 & 2  & 1.638\\
    \hline
       \multirow{3}{*}{Two-stream PDM}  & 11.29 & 27.08 & 1  & 1.522\\
       &  10.99& 27.53 & 2 & 1.592\\
        & 30.91 & 25.87 &4   & 1.638\\
        \hline
    \end{tabular}
    \caption{The effect of segmentation patch size on FID, CLIP, and MCD of generated images and masks}
    \label{tab:patch}
\end{table}

\subsection{Quantitative Evaluation}
We evaluate the quality of generated images using FID \cite{NIPS2017_FID} and CLIP scores \cite{hessel2022clipscore}. FID assesses the quality and fidelity of the generated images by employing an Inception model, while CLIP scores gauge how well the generated images correspond to the text prompts. For CLIP scores, we use the ViT-B/32 model \cite{radford2021clip}. We generate 30,000 images and segmentation maps from 5,000 text files in the COCO dataset's validation set, with each file containing five captions describing the same scene. We compute the average CLIP scores by comparing five captions with the generated images.

In Table.\ref{tab:results}, we compare the FID and CLIP scores of our models with those of state-of-the-art methods. The results indicate that while our panoptic diffusion models (PDMs) are trained with a combined loss of images and segmentation maps, they achieve comparable fidelity (FID scores) and improved relevance between image and text (higher CLIP scores). This improvement is due to the enhanced connectivity between the image, text, and segmentation map. The two-stream PDM performs better due to its pretrained stream and larger number of parameters. When ground-truth maps are provided, the model performs optimally because it focuses solely on optimizing image generation.

Table \ref{tab:patch} shows increasing the patch factor results in a higher MCD because generating higher-resolution maps with a fixed number of latents becomes more challenging. This creates a trade-off between map resolution and quality. We find that a patch factor of 2 offers the best balance, yielding the highest FID and CLIP. However, increasing the patch factor to 4 results in worse performance, suggesting that unbalanced patch sizes for maps and images are detrimental.

Please see Appendix. \ref{appen.ablation} for more ablation study.

\begin{figure}[tbp]
\begin{subfigure}[h]{0.45\linewidth}
\includegraphics[width=\linewidth]{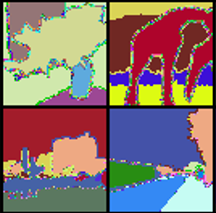}
\caption{Ground-truth segmentation maps}
\end{subfigure}
\hfill
\begin{subfigure}[h]{0.45\linewidth}
\includegraphics[width=\linewidth]{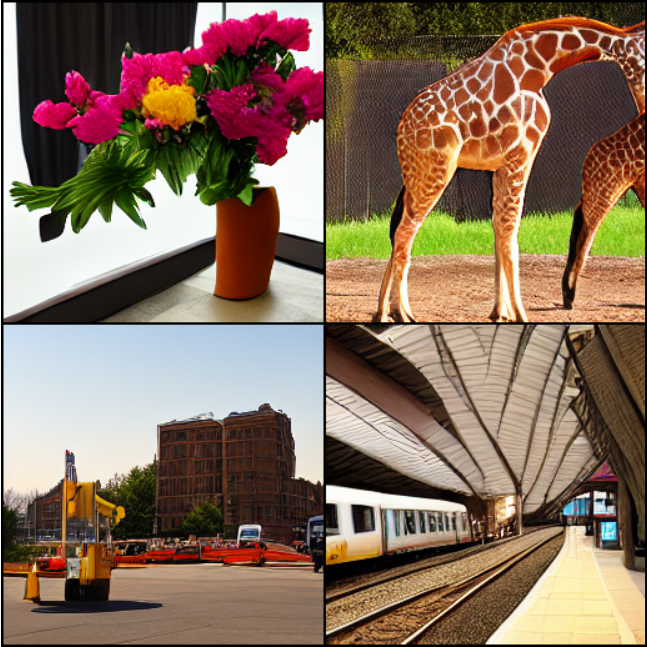}
\caption{Images generated based on ground-truth maps}
\end{subfigure}%

\begin{subfigure}[h]{0.45\linewidth}
\includegraphics[width=\linewidth]{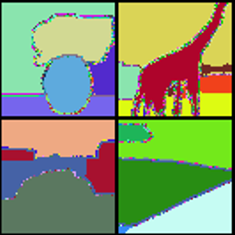}
\caption{Generated maps}
\end{subfigure}
\hfill
\begin{subfigure}[h]{0.45\linewidth}
\includegraphics[width=\linewidth]{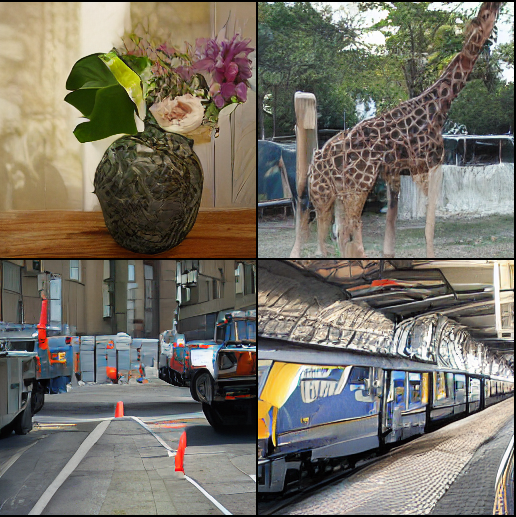}
\caption{Images co-generated with maps}
\end{subfigure}%
\caption{Image-map co-generation. Prompts are: 1) a small copper vase with some flowers in it; 2) A giraffe examining the back of another giraffe; 3) A utility truck is parked in the street beside traffic cones; 4) A white yellow and blue train at an empty train station.}\label{fig.cogeneration}
\end{figure} 

\subsection{Qualitative Evaluaiton}
In Fig. \ref{fig:compare}, we compare the images and masks generated by PDM with images generated by U-ViT. By training with segmentation masks, PDM learns that the shape of a stop sign should be octagon, while U-ViT cannot guarantee to generate an octagon stop sign. Similarly, PDM ensures to generate correct shapes for a fire hydrant and a human. In the last row of Fig. \ref{fig:compare}, PDM generates masks for not only elephants but also for the river, while a regular diffusion model misses the required component of the text prompt.\\
Figure \ref{fig.cogeneration} displays images generated with either ground-truth segmentation maps or co-generated maps. The generated maps in the bottom left show objects of the same categories and similar shapes as the ground-truth maps. The images on the right are conditioned on these segmentation maps, demonstrating the PDM's ability to generate correlated images and maps. While images generated with ground-truth maps exhibit slightly better quality, co-generation removes the need for a segmentation input and produces diverse maps and images. Additional examples generated by PDMs are provided in Appendix \ref{appen.images}. The color map of categories are shown in Appendix \ref{app.color}.


\section{Conclusion}
In conclusion, we introduce the Panoptic Diffusion Model (PDM), a pioneering approach that simultaneously generates images and panoptic segmentation maps from a given prompt. Unlike previous diffusion models that either depend on pre-existing segmentation maps or generate them based on images, PDM inherently understands and constructs scene layouts during the generation process. This innovation enables PDM to produce more creative and realistic images by leveraging segmentation layouts as intrinsic guidance. This research lays the groundwork for future advancements in diffusion models, offering a robust framework for co-generation of images and segmentation maps.

\bibliography{example_paper}
\bibliographystyle{icml2025}

\newpage
\appendix
\onecolumn
\section{Fast DPM solver for segmentation maps} \label{appen.dpm}
We modify the first order and third order DPM-solver++ to solve the image and map of the previous step given $x_t$, $M_t$ and predicted $x_0$, $M_0$ \cite{lu2023dpmsolverfastsolverguided}.  The pseudo code for the solvers are listed below. For the details of the algorithm and definition of the parameters $\sigma, \alpha, \phi, s$, please check DPM-solver++.
\begin{lstlisting}
def dpmFirstSolver(self,x_0, m_0, x_t,m_t):
    x_t=(sigma_t/sigma_s)*x+(alpha_t*phi_1)*x_0
    #update  M[t-1] based on M[t]
    m_t= (sigma_t/sigma_s)*m_t +
        (alpha_t*phi_1)*m_0
    return x_t, m_t

def dpmThirdSolver(self, x_t,m_t,C,t):
    #First step
    x_0, m_0= diffusionModel(x_t,m_t,C,s)
    x_s1=(sigma_s1/sigma_s)*x+(alpha_s1*phi_11)*x_0
    m_s1= (sigma_s1/sigma_s)*m_t +
        (alpha_s1*phi_11)*m_0
    #Second step
    x_02, m_02= diffusionModel(x_s1,m_s1,C,s1)
    x_s2=(sigma_s2/sigma_s)*x+(alpha_s1*phi_12)*x_0 + 
    r2 / r1 * (alpha_s2 * phi_22)* (x_02 - x_0)
    m_s2= (sigma_s2/sigma_s)*m_t +
        (alpha_s2*phi_12)*m_0 + 
    r2 / r1 * (alpha_s2 * phi_22)* (m_02 - m_0)
    #Third step
    x_03, m_03= diffusionModel(x_s2,m_s2,C,s2)
    x_t=(sigma_t/sigma_s)*x+(alpha_t*phi_1)*x_0 + 
   (1. / r2) * (alpha_t * phi_2)* (x_03 - x_0)
    m_t= (sigma_t/sigma_s)*m_t +
        (alpha_t*phi_1)*m_0 + 
   (1. / r2) * (alpha_t * phi_2)* (m_03 - m_0)
    return x_t, m_t
\end{lstlisting}
\section{Ablation study}
\label{appen.ablation}

\begin{figure}[htbp]
\begin{subfigure}[h]{0.4\linewidth}
\centering
\includegraphics[width=\linewidth]{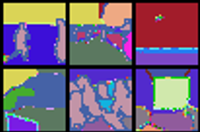}
\caption{32x32 maps if $\epsilon_M$ is $\mathcal{N}(0,2*\mathbf{I})$}
\end{subfigure}
\hfill
\begin{subfigure}[h]{0.4\linewidth}
    \centering
    \includegraphics[width=\linewidth]{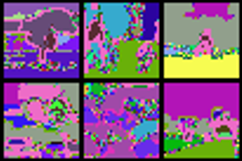}
    \caption{32x32 maps if $\epsilon_M$ is $\mathcal{N}(0,\mathbf{I})$}
    \label{fig.noise}
\end{subfigure}
\hfill
\begin{subfigure}[h]{0.4\linewidth}
\centering
\includegraphics[width=\linewidth]{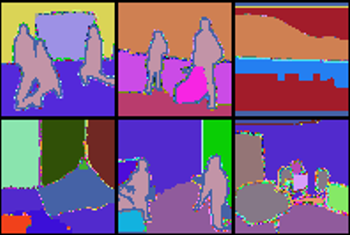}
\caption{64x64 maps}
\end{subfigure}
\hfill
\begin{subfigure}[h]{0.4\linewidth}
\centering
\includegraphics[width=\linewidth]{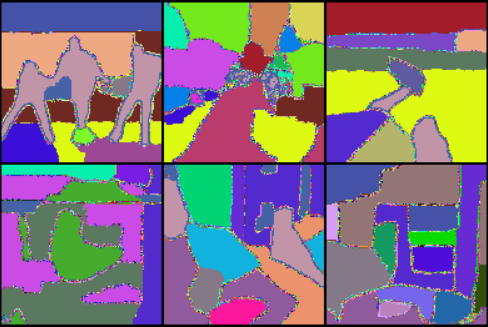}
\caption{128x128 maps}
\end{subfigure}%

\caption{Generated maps of different resolutions. Prompts are 1)Three people are playing with a red kick ball; 2) A woman walking next to a man riding a pink bike; 3) An old man is flying his kite in the middle of no where; 4) A large lizard sitting on stone steps with three birds;
5) A girl is playing a game system while other kids look on; 6) A living room that has some couches and tables in it} \label{fig.resolution}
\end{figure}
 \subsection{Effect of the patch factor}  We evaluate the impact of different patch sizes on map resolution, as illustrated in Figure \ref{fig.resolution}. When the patch size for segmentation maps is set to four times that of the images, the resulting maps have a resolution of 128x128. However, these larger maps may include hallucinated details that could misguide image generation. This issue arises due to the disparity in patch sizes and the model's limited hidden dimension of 768, which complicates accurate prediction for a 128x128 map.

\subsection{Replacing noisy map inputs with zero}
To assess whether PDMs learn to denoise the segmentation map or extract it from the image latent, we replace noisy map inputs $M_t$ with zero inputs during training. The results reveals that while a two-stream model can still generate images (FID=18.94), it cannot generate readable maps. This indicates that a panoptic diffusion model does not solely depend on image features for map generation, unlike the approach in DiffuMask \cite{wu2024diffumasksynthesizingimagespixellevel}. Hence, noisy map inputs $M_t$ are crucial for predicting $M_0$.
\subsection{Noise scale for segmentation maps} As previously mentioned, the noise added to segmentation maps must be greater than one to effectively flip the analog bits. If the noise variance is smaller than one, it fails to convert the training signal to noise at any timestep, resulting in the model's inability to denoise maps adequately. Figure \ref{fig.noise} demonstrates that maps are not properly denoised when $\epsilon_M\sim\mathcal{N}(0,\mathbf{I})$.

\section{More examples of generated images and maps \label{appen.images}}
\subsection{Comparison between using ground-truth segmentation map and using co-generated maps}

\begin{figure}[htbp]
\begin{subfigure}[h]{0.8\linewidth}
\includegraphics[width=\linewidth]{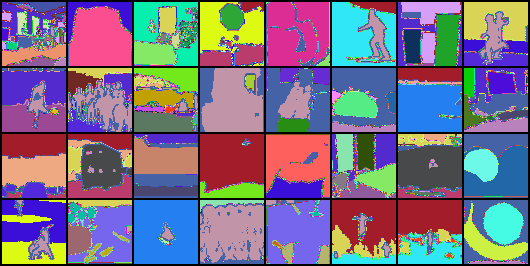}
\caption{Ground-truth segmentation maps}
\end{subfigure}
\hfill
\begin{subfigure}[h]{0.8\linewidth}
\includegraphics[width=\linewidth]{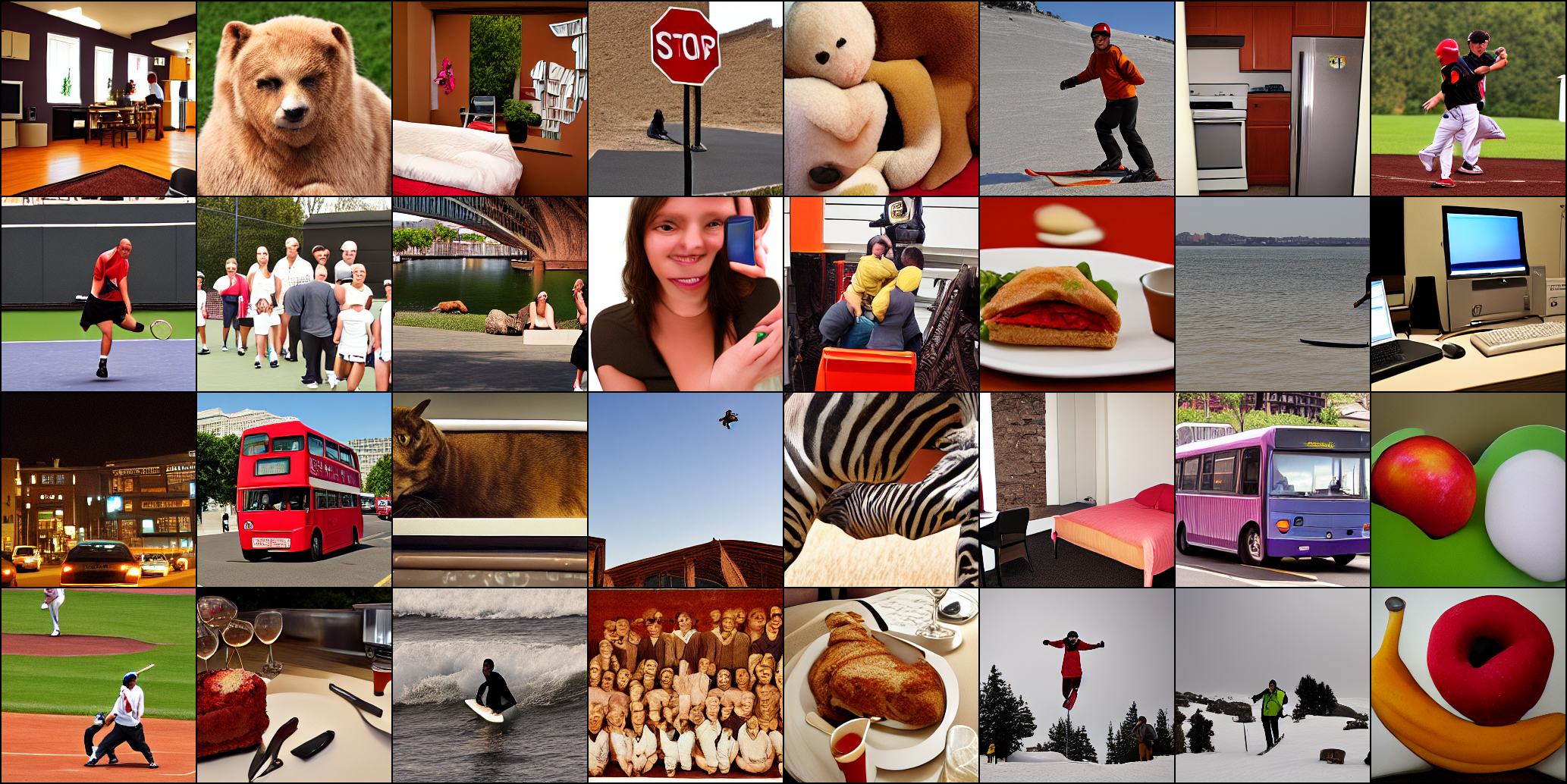}
\caption{Generated images based on ground-truth maps}
\end{subfigure}%
\caption{Generation with given segmentation maps } \label{moresamples}
\end{figure}

\begin{figure}
\begin{subfigure}[h]{0.8\linewidth}
\includegraphics[width=\linewidth]{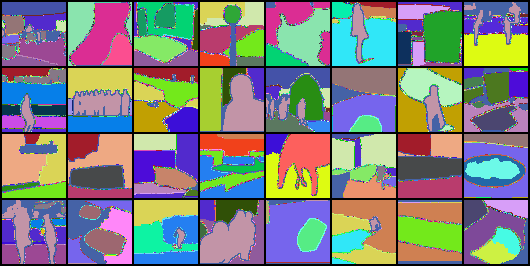}
\caption{Generated segmentation naps}
\end{subfigure}%
\hfill
\begin{subfigure}[h]{0.8\linewidth}
\includegraphics[width=\linewidth]{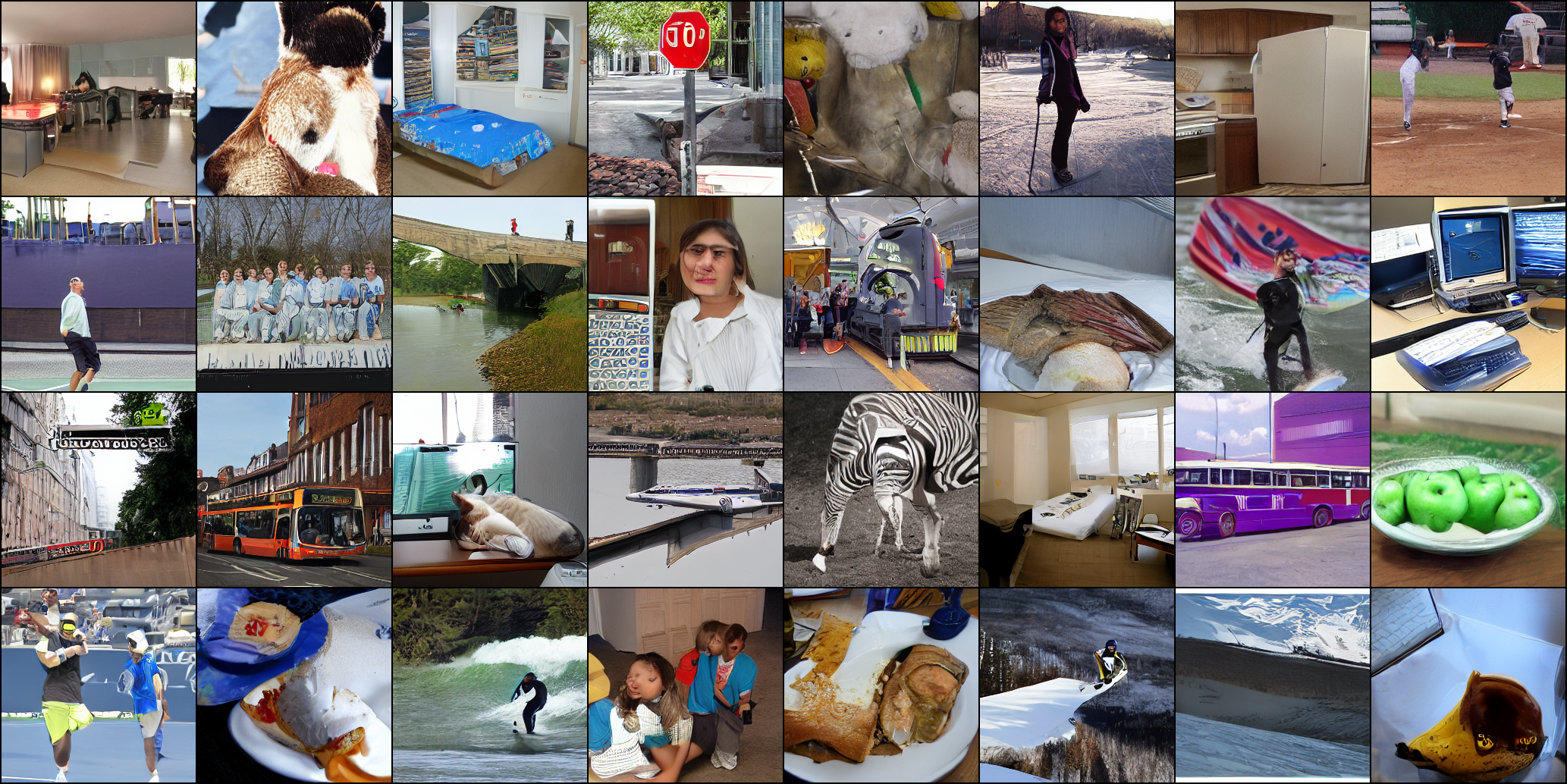}
\caption{Co-generated images}
\end{subfigure}%
\hfill
\begin{subfigure}[h]{0.8\linewidth}
\includegraphics[width=\linewidth]{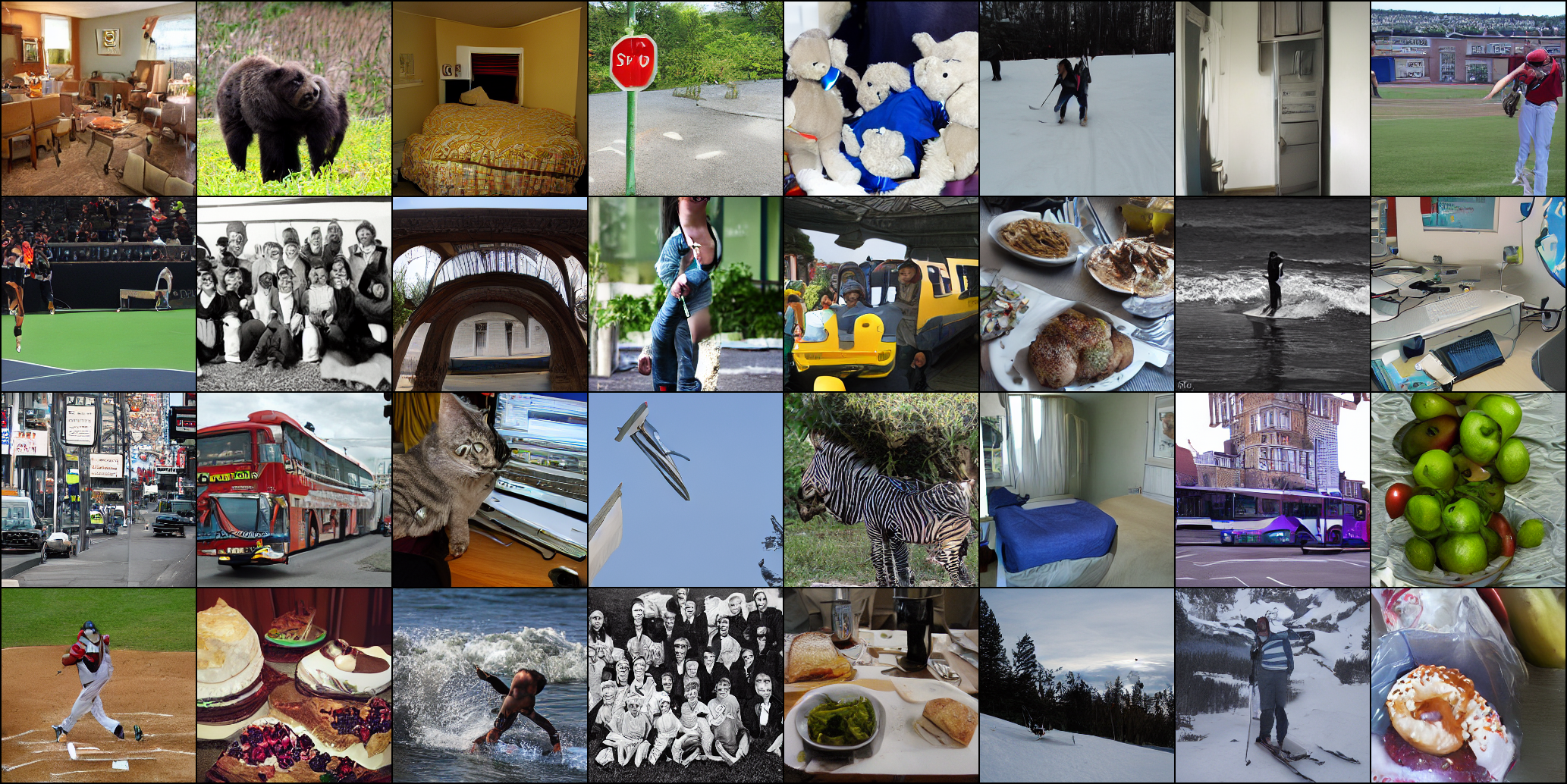}
\caption{Images generated by U-ViT (baseline)}
\end{subfigure}%
\caption{Cogeneration of images and segmentation maps} 
\end{figure}
Fig. \ref{moresamples} shows more examples of generated images and segmentation maps. The prompts are randomly chosen from COCO2017 validation dataset, as listed below.\\
 0 A woman stands in the dining area at the table.\\
 1 A big burly grizzly bear is show with grass in the background.\\
 2 Bedroom scene with a bookcase, blue comforter and window.\\
 3 A stop sign is mounted upside-down on it's post. \\
 4 Three teddy bears, each a different color, snuggling together.\\
 5 A woman posing for the camera standing on skis.\\
 6 A kitchen with a refrigerator, stove and oven with cabinets.\\
 7 A couple of baseball player standing on a field.\\
 8 a male tennis player in white shorts is playing tennis\\
 9 The people are posing for a group photo.\\
 10 A beautiful woman taking a picture with her smart phone.\\
 11A woman holding a Hello Kitty phone on her hands.\\
 12some children are riding on a mini orange train\\
 13A meal is lying on a plate on a table.\\
 14A man in a wet suit stands on a surfboard and rows with a paddle.\\
 15A computer on a desk next to a laptop.\\
 16A street scene with focus on the street signs on an overpass.\\
 17The red, double decker bus is driving past other buses. \\
 18A cat resting on an open laptop computer.\\
 19Two planes flying in the sky over a bridge.\\
 20A zebra in the grass who is cleaning himself. \\
 21A bedroom with a bed and small table near by.\\
 22a big purple bus parked in a parking spot\\
 23A large white bowl of many green apples. \\
 24Batter preparing to swing at pitch during major game.\\
 25A plate of finger foods next to a blue and raspberry topped cake.\\
 26A man on a blue raft attempting to catch a ride on a large wave.\\
 27Many small children are posing together in the black and white photo. \\
 28A plate on a wooden table full of bread.\\
 29A man flying through the air while riding skis.\\
 30A person standing on top of a ski covered slope.\\
 31a close up of a banana and a doughnut in a plastic bag

\begin{figure}[htbp]
\begin{subfigure}[h]{1\linewidth}
\centering
\includegraphics[width=0.7\linewidth]{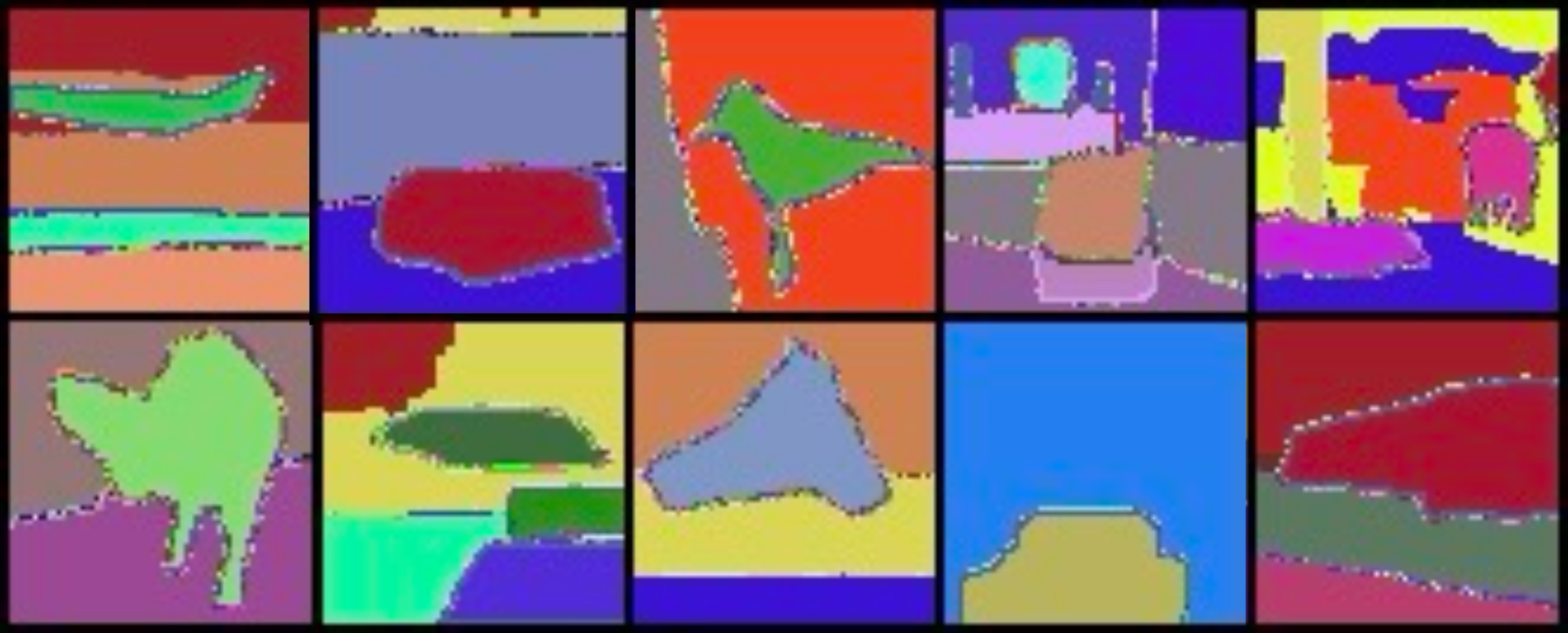}
\caption{Generated Segmentation Map}
\end{subfigure}
\hfill
\begin{subfigure}[h]{1\linewidth}
\centering
\includegraphics[width=0.7\linewidth]{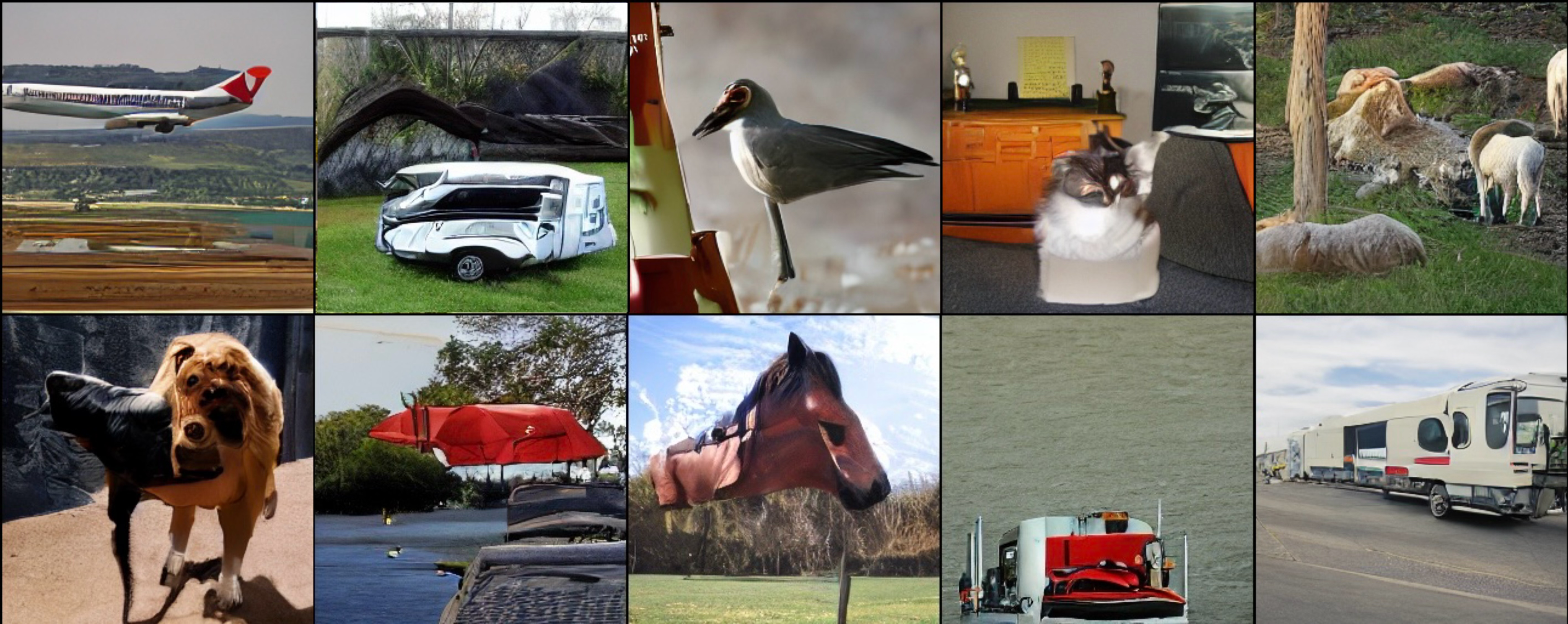}
\caption{Generated Image}
\end{subfigure}%
\caption{Zero-shot evaluation on CIFAR10}
\end{figure}

\begin{figure*}[hbtp]
    \centering
    \includegraphics[width=0.9\linewidth]{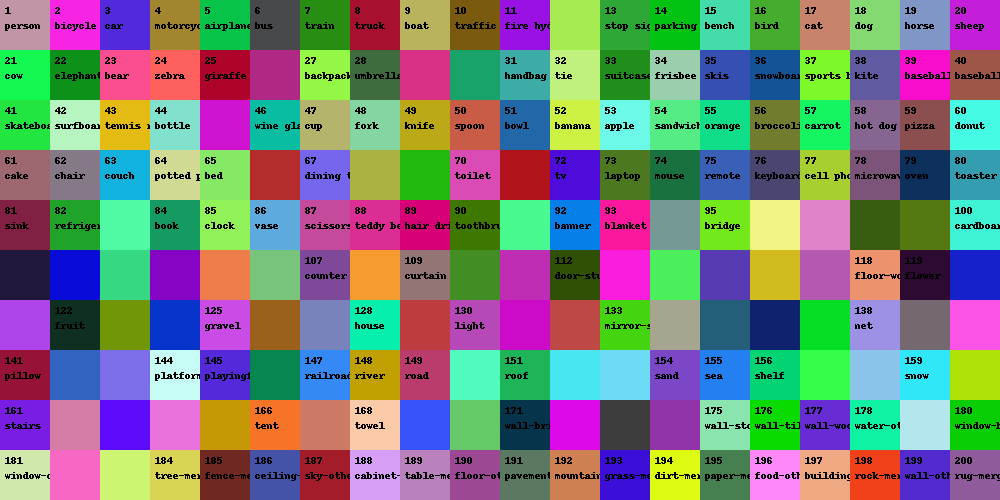}
    \caption{Color map}
    \label{fig:color}
\end{figure*}

\subsection{Zero-shot results on CIFAR10}
We apply the model trained on COCO dataset to generate images with segmentation maps for CIFAR10. The class labels are encoded by the text encoder as image captions. The zero-shot results show that our model is capable of generating segmentation maps for things and stuffs for other image datasets.


\section{Color map of panoptic categories of COCO dataset}
\label{app.color}
The pixel values in the generated segmentation maps correspond to category IDs (1-200), which are mapped to random RGB colors for visualization. Please see Fig. \ref{fig:color}. This is a random color map only for reference. Although COCO dataset uses 1-200 as class labels, there are only 133 classes.


\end{document}